\documentclass[runningheads]{llncs}
\usepackage{graphicx}
\usepackage{amsmath,amssymb,amsfonts}
\usepackage{multicol} 
\usepackage{caption} 
\usepackage{wrapfig} 
\def\state{s}
\def\stateSpace{\mathcal{S}}
\def\ctrl{a}

\def\Cparam{\nu}
\def\Aparam{\theta}
 
\def\real{\mathbb R}
\def\natural{\mathbb N}

\def\comment#1{}

\def\eqref#1{(\ref{#1})}
\def\Beq#1\Eeq{\begin{equation}#1\end{equation}}
\def\Beqo#1\Eeqo{\begin{equation*}#1\end{equation*}}
\def\Beqs#1\Eeqs{\begin{align}#1\end{align}}
\def\Beqso#1\Eeqso{\begin{align*}#1\end{align*}}

\begin{document}

\title{A framework for reinforcement learning \\ with autocorrelated actions} 
%
\author{Marcin Szulc\inst{1} \and
Jakub \L{}yskawa\inst{1} \and
Pawe\l{} Wawrzy\'nski\inst{1}\orcidID{0000-0002-1154-0470}}

\authorrunning{M. Szulc, J. \L{}yskawa, P. Wawrzy\'nski}
%
\institute{Warsaw University of Technology, Institute of Computer Science, Warsaw, Poland \\
\email{\{marcin.szulc,jakub.lyskawa\}.stud@pw.edu.pl}, \email{pawel.wawrzynski@pw.edu.pl}}
\maketitle              
\begin{abstract}
The subject of this paper is reinforcement learning. Policies are considered here that produce actions based on states and random elements autocorrelated in subsequent time instants. Consequently, an~agent learns from experiments that are distributed over time and potentially give better clues to policy improvement. Also, physical implementation of such policies, e.g. in robotics, is less problematic, as it avoids making robots shake. This is in opposition to most RL algorithms which add white noise to control causing unwanted shaking of the robots. 
An~algorithm is introduced here that approximately optimizes the aforementioned policy. Its efficiency is verified for four simulated learning control problems (Ant, HalfCheetah, Hopper, and Walker2D) against three other methods (PPO, SAC, ACER). The algorithm outperforms others in three of these problems. 

\end{abstract}
%
%

\section{Introduction} 

The usual goal of Reinforcement Learning (RL) to optimize a~policy that samples an action on the basis of a~current state of a~learning agent. The only stochastic dependence between subsequent actions is through state transition: The action moves the agent to another state which determines the distribution of another action. Main analytical tools in RL are based on this lack of other dependence between actions.  E.g., for a~given policy, its value function expresses the expected sum of discounted rewards the agent may expect starting from a~given state. The sum of rewards does not depend on actions taken before the given state was reached. Hence, only the given state and the policy matter. 

Lack of dependence between actions beyond state transition leads to several difficulties. In physical implementation of RL, e.g. in robotics, it usually means that white noise is added to control actions. However, that makes control discontinuous and rapidly changing all the time. This is often impossible to implement since electric motors that are to execute these actions can not operate this way. Even if it is possible, it requires a~lot of energy, makes the controlled system shake, and exposes it to damages. 

It is also questionable if the lack of dependence between actions beyond state transition does not reduce efficiency of learning. Each action is an~experiment that leads to policy improvement. However, due to limited accuracy of (act\mbox{ion-)}value function approximation, consequences of a~single action may be difficult to recognize. The finer the time discretization, the more serious this problem becomes. Consequences of a~random experiment distributed over several time instants could be more tangible thus easier to recognize. 

The contribution of this paper may be summarized in the following points: 
\begin{itemize} 
\item 
A framework is introduced in which a~policy produces actions on the basis of states and values of a~stochastic process. That enables relation between actions that is beyond state transition. 
\item 
An algorithm is introduced that approximately optimizes the aforementioned policy. 
\item 
The above algorithm is tested on four benchmark learning control problems: Ant, Half-Cheetah, Hopper, and Walker2D. 
\end{itemize} 

The rest of the paper is organized as follows. Section~\ref{sec:related-work} overviews related literature. Sec.~\ref{sec:policy} introduces a~policy that produces autocorrelated actions along with tools for its analysis. Sec.~\ref{sec:alg} introduces an~algorithm that approximately optimizes that policy. Sec.~\ref{sec:experiments} presents simulations that compare the presented algorithm with state-of-the-art reinforcement learning methods. The last section concludes the paper. 

\section{Related Work} 
\label{sec:related-work} 

\subsection{Stochastic dependence between actions} 

The idea of introducing stochastic dependence between actions was analyzed in~\cite{2007wawrzynski} as a~remedy to problems with application of RL in fine time discretization. The control process was divided there into ``non-Markov periods'' in which actions were stochastically dependent. 
A~policy with autocorrelated actions was analyzed in \cite{2015wawrzynski} with a~standard RL algorithm applied to its optimization that did not account for the dependence of actions. 

In~\cite{2017vanhoof+2} a~policy was analyzed whose parameters were incremented by the~autoregressive stochastic process. Essentially, this resulted in autocorrelated random components of actions. In~\cite{2019korenkevych+3} a~policy was analyzed that produced an~action being a~sum of the~autoregressive noise and a~deterministic function of state. However, no learning algorithm was presented in this paper that accounted for specific properties of this policy. 

\subsection{Reinforcement learning with experience replay} 

The Actor-Critic architecture of reinforcement learning was introduced in~\cite{1983barto+2}. Approximators were applied to this structure for the first time in~\cite{1998kimura+2}. In order to boost efficiency of these algorithms, they were combined with experience replay for the first time in~\cite{2009wawrzynski}. 

Application of experience replay to Actor-Critic encounters the following problem. The learning algorithm needs to estimate quality of a~given policy on the basis of consequences of actions that were registered when a~different policy was in use. Importance sampling estimators are designed to do that, but they can have arbitrarily large variance. In~\cite{2009wawrzynski} that problem was addressed with truncating density ratios present in those estimators. In~\cite{2016wang+6} specific correction terms were introduced for that purpose. 

Another approach to the aforementioned problem is to prevent the algorithm from inducing a~policy that differs too much from the one tried. That idea was first applied in Conservative Policy Iteration~\cite{2002kakade+1}. It was further extended in Trust Region Policy Optimization~\cite{2015schulman+4}. This algorithm optimizes a~policy with the constraint that the Kullback-Leibler divergence between that policy and the tried one should not exceed a~given threshold. The K-L divergence becomes an~additive penalty in Proximal Policy Optimization algorithms, namely PPO-Penalty and PPO-Clip~\cite{2017schulman+4}. 

A~way to avoid the problem of estimating quality of a~given policy on the basis of the tried one is to approximate the action-value function instead of estimating the value function. Algorithms based on this approach are Deep Q-Network (DQN)~\cite{2013mnih+6}, Deep Deterministic Policy Gradient (DDPG)~\cite{2016lillicrap+7}, and Soft Actor-Critic (SAC)~\cite{2018haarnoja+3}. In the original version of DDPG the time-correlated OU noise was added to action. However, this algorithm was not adapted to this fact in any specific way. SAC uses white noise in actions and it is considered one of the most efficient in this family of algorithms. 

\section{Policy with autocorrelated actions} 
\label{sec:policy} 

Let an~action, $\ctrl_t$, be computed as 
\Beq \label{def:pi} 
    \ctrl_t = \pi(\state_t,\xi_t;\Aparam) 
\Eeq
where $\pi$ is a~deterministic transformation, $\state_t$ is a~current state, $\Aparam$ is a~vector of trained parameters, and $(\xi_t,t=1,2,\dots)$ is a~stochastic process. We require this process to have the following properties: 
\begin{itemize} 
\item 
Stationarity: The distribution of $\xi_t$ is the same for each $t$. 
\item 
Zero mean: $E\xi_t = 0$ for each $t$. 
\item 
Autocorrelation decreasing with growing lag: 
\Beq
    E\xi_t^T\xi_{t+k} > E\xi_t^T\xi_{t+k+1} \geq 0 
    \;\text{for}\; k\geq0. 
\Eeq
Essentially that means that values of the process are close to each other when they are in close time instants. 
\item 
Markov property: For any $t$ and $k,l\geq0$, the conditional distributions 
\Beq \label{xi:markov:prop} 
    (\xi_{t},\dots,\xi_{t+k} | \xi_{t-1}, \dots, \xi_{t-1-l}) 
    \quad \text{and} \quad
    (\xi_{t},\dots,\xi_{t+k} | \xi_{t-1})
\Eeq
are the same. In words, dependence of future values of $(\xi_t)$ on its past is entirely carried over by $\xi_{t-1}$.
\end{itemize} 

Consequently, if only $\pi$ \eqref{def:pi} is continuous for all its arguments, and subsequent states $\state_t$ are close to each other, then the corresponding actions are close, although random. In words, they create a~consistent, distributed in time experiment that can lead to policy improvement. 

\begin{wrapfigure}{r}{0.5\textwidth}
  \vspace{-3em}
  \begin{center}
    \includegraphics[width=0.48\textwidth]{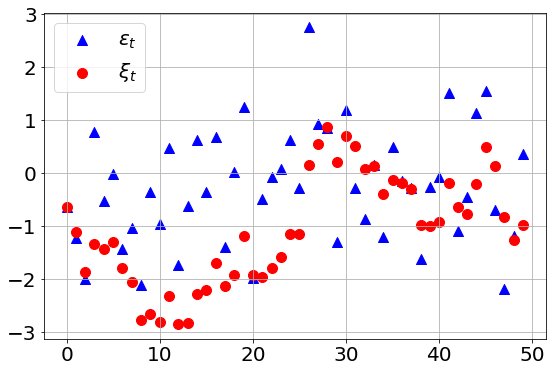}
  \end{center}
  \vspace{-1em}
  \caption{A realization of the~normal white noise $(\epsilon_t)$, and the auto-regressive process $(\xi_t)$ \eqref{AR:xi}. }
  \label{fig:xi} 
  \vspace{-4em}
\end{wrapfigure}

\paragraph{Example: Auto-Regressive $(\xi_t)$.} Let $\alpha \in [0,1)$ and 
\Beq \label{AR:xi} 
    \begin{split} 
    \epsilon_t & \sim N(0,C), \quad t=1,2,\dots \\ 
    \xi_1 & = \epsilon_1 \\ 
    \xi_t & = \alpha \xi_{t-1} + \sqrt{1-\alpha^2} \epsilon_t, \quad t=2,3,\dots 
    \end{split} 
\Eeq
Fig.~\ref{fig:xi} demonstrates a~realization of both the white noise $(\epsilon_t)$ and $(\xi_t)$. Let us analyze if $(\xi_t)$ has the required properties. 

Both $\epsilon_t$ and $\xi_t$ have the same distribution $N(0,C)$. Therefore $(\xi_t)$ is stationary and zero-mean. 
A simple derivation reveals that 
$$
    E\xi_{t}\xi_{t+k}^T = \alpha^{|k|} C 
    \; \text{ and } \;
    E\xi_{t}^T\xi_{t+k} = \alpha^{|k|} \text{tr}(C) 
$$ 
for any $t,k$. Therefore, $(\xi_t)$ is autocorrelated, and this autocorrelation decreases with growing lag. Consequently, the values of $\xi_t$ are closer to one another for subsequent $t$ than the values of $\epsilon_t$, namely 
\Beqso
    E\|\epsilon_{t} - \epsilon_{t-1}\|^2
    & = E(\epsilon_{t} - \epsilon_{t-1})^T(\epsilon_{t} - \epsilon_{t-1}) 
    = 2\text{tr}(C) \\ 
    E\|\xi_{t} - \xi_{t-1}\|^2 
    & = E\left((\alpha\!-\!1)\xi_{t-1} + \sqrt{1-\alpha^2}\epsilon_t\right)^T\left((\alpha-1)\xi_{t-1} + \sqrt{1-\alpha^2}\epsilon_t\right) \\ 
    & = (\alpha-1)^2 \text{tr}(C) + (1-\alpha^2) \text{tr}(C) = (1-\alpha)2\text{tr}(C). 
\Eeqso

The Markov property of $(\xi_t)$ directly results from how $\xi_t$~\eqref{AR:xi} is computed. 

In fact, marginal distributions of the process $(\xi_t)$, as well as its conditional marginal distributions are normal, and their parameters have compact forms. We shall not present derivations of these parameters due to lack of space, but we shall denote them for further use. Namely, let as consider 
\Beq
    \bar\xi^n_t = [\xi_t^T, \dots, \xi_{t+n-1}^T]^T. 
\Eeq
The distribution of $\bar\xi^n_t$ is normal 
\Beq \label{Omega_0} 
    N(0, \Omega^n_0), 
\Eeq
where $\Omega^n_0$ is a~matrix dependent on $n$, $\alpha$, and $C$. The conditional distribution $(\bar\xi^n_t|\xi_{t-1})$ is also normal, 
\Beq \label{B,Omega_1}
    N(B^n\xi_{t-1}, \Omega^n_1), 
\Eeq
where both $B^n$ and $\Omega^n_1$ are matrices dependent on $n$, $\alpha$, and $C$.

\paragraph{The neural-normal policy.} A simple and practical way to implement $\pi$ \eqref{def:pi} is as follows. A~feedforward neural network, 
\Beq
    A(\state;\Aparam), 
\Eeq
has input $\state$ and weights $\Aparam$. An~action is computed as 
\Beq \label{ctrl=A+xi} 
    \ctrl_t = \pi(\state_t,\xi_t;\Aparam) = A(\state_t;\Aparam) + \xi_t,  
\Eeq
for $\xi_t$ in the form~\eqref{AR:xi}. 
While the discussion below can be extended to the general formulation~\eqref{def:pi}, in order to make it simpler we will further assume that a~policy is of the form~\eqref{ctrl=A+xi}. 

Let us consider 
\Beqso
    \bar\state^n_t & = [\state_{t}^T,\dots,\state_{t+n-1}^T]^T, \\ 
    \bar\ctrl^n_i & = [\ctrl_{t}^T,\dots,\ctrl_{t+n-1}^T]^T, \\ 
    \bar A(\bar\state^n_i;\Aparam) & = 
    [A(\state_{t};\Aparam)^T,\dots,A(\state_{t+n-1};\Aparam)^T]^T,  
\Eeqso
and fixed $\Aparam$. With \eqref{ctrl=A+xi} the distributions $(\bar\ctrl^n_t|\bar\state^n_t)$ and $(\bar\ctrl^n_t|\bar\state^n_t, \xi_{t-1})$ are both normal, namely $N(\bar A(\bar\state^n_t;\Aparam), \Omega^n_0)$, and $N(\bar A(\bar\state^n_t;\Aparam) + B^n\xi_{t-1}, \Omega^n_1)$, respectively (see \eqref{Omega_0} and \eqref{B,Omega_1}). The algorithm defined in the next section updates $\Aparam$ to manipulate the above distributions. Density of the normal distribution with mean $\mu$ and covariance matrix $\Omega$ will be denoted by 
\Beq
    \varphi(\cdot\,;\mu,\Omega). 
\Eeq

\paragraph{Noise-value function.} In policy \eqref{def:pi} there is a~stochastic dependence between actions beyond the dependence resulting from state transition. Therefore, the traditional understanding of policy as distribution of actions conditioned on state does not hold here. Each action depends on the current state, but also previous states and actions. Analytical usefulness of the traditional value function and action-value function is thus limited. 

As a valid analytical tool we propose {\it noise-value function} defined as \Beq \label{def:W} 
    W^\pi(\xi,\state) 
    = E_\pi\left(\sum_{i\geq0} \gamma^i r_{t+i} \Big| \xi_{t-1} = \xi, \state_t = \state\right). 
\Eeq
The course of events starting in time $t$ depends on the current state $\state_t$ and the value $\xi_{t-1}$. Because of Markov property of $\xi_t$ \eqref{xi:markov:prop}, the pair $(\xi_{t-1},\state_t)$ is a~proper condition for the expected value of future rewards. 

The {\it value function} $V^\pi : \stateSpace \mapsto \real$ is slightly redefined, namely 
\Beq \label{def:V} 
    V^\pi(\state) = E\big(W(\xi_{t-1},\state_t) | \state_t = \state\big). 
\Eeq
The random value in the above expectation is $\xi_{t-1}$ and its distribution is conditional with the condition $\state_t = \state$. The distribution of $\xi_{t-1}$ may differ for different~$\state_t$. However, being in the state $\state_t$ and not knowing $\xi_{t-1}$ the agent may expect the sum of future rewards equal to~$V^\pi(\state_t)$.

\section{ACERAC: Actor-Critic with Experience Replay and Autocorrelated aCtions} 
\label{sec:alg} 

The algorithm presented here has Actor-Critic structure. It optimizes a~policy of the form~\eqref{ctrl=A+xi} and uses Critic, 
$$
    V(\state;\Cparam), 
$$
which is an approximator of the value function~\eqref{def:V} parametrized by a~vector,~$\Cparam$. 

For each time instant of the~agent-environment interaction the policy~\eqref{ctrl=A+xi} is applied and a~tuple, $\langle \state_t, A_t, \ctrl_t, r_t, \state_{t+1} \rangle$, is registered, where $A_t = A(\state_t;\Aparam)$. 

The general goal of training is to maximize $W^\pi(\xi_{i-1},\state_i)$ for each state $\state_i$ registered during the agent-environment interaction. In this order previous time instants are sampled, and sequences of actions that follow these instants are made more/less probable depending on their return. More specifically, $i$ is sampled from $\{1,\dots,t-1\}$ and the conditional density of the sequence of actions $(\ctrl_i, \dots, \ctrl_{i+n-1})$ is being increased/dec\-reas\-ed depending on the return 
$$
    r_i + \dots + \gamma^{n-1} r_{i+n-1} + \gamma^{n} V(\state_{i+n};\Cparam) 
$$
this sequence of actions yields. 
At the same time adjustments of the same form are performed for several sequences of actions starting from $\ctrl_i$, namely for $n=1,\dots,\tau$, where $\tau \in \natural$ is a~parameter. 

\subsection{Actor \& Critic training} 

The following procedure is repeated several times at each $t$-th instant of agent--environment interaction: 
\begin{enumerate} 
\item A~random $i$ is sampled from the uniform distribution over $\{1, \dots, t-1\}$.
\item If $i$ is the initial instant of a~trial, then consider for $n=1,\dots,\tau$
\Beqso
    \mu_{i+j} & = E(\xi_{i+j}) = 0, \; j=0,\dots,n-1 \\ 
    \eta_{i+j} & = E(\xi_{i+j}) = 0, \; j=0,\dots,n-1 \\ 
    \Omega^n_2 & = \Omega^n_0. 
\Eeqso
Otherwise, consider 
\Beqso
    \mu_{i+j} & = E(\xi_{i+j}|\xi_{i-1} = \ctrl_{i-1} - A_{i-1}), \; j=0,\dots,n-1 \\ 
    \eta_{i+j} & = E(\xi_{i+j}|\xi_{i-1} = \ctrl_{i-1} - A(\state_{i-1};\Aparam)), \; j=0,\dots,n-1 \\ 
    \Omega^n_2 & = \Omega^n_1. 
\Eeqso
\item Consider the following vectors for $n=1,\dots,\tau$
\Beqso
    \bar\mu^n_i & = [\mu_{i}^T,\dots,\mu_{i+n-1}^T]^T, \\ 
    \bar\eta^n_i & = [\eta_i^T,\dots,\eta_{i+n-1}^T]^T, \\ 
    \bar\state^n_i & = [\state_{i}^T,\dots,\state_{i+n-1}^T]^T, \\ 
    \bar\ctrl^n_i & = [\ctrl_{i}^T,\dots,\ctrl_{i+n-1}^T]^T, \\ 
    \bar A^n_i & = [A_{i}^T,\dots,A_{i+n-1}^T]^T, \\
    \bar A(\bar\state^n_i;\Aparam) & = 
    [A(\state_{i};\Aparam)^T,\dots,A(\state_{i+n-1};\Aparam)^T]^T. 
\Eeqso
\item Temporal differences are computed for $n=1,\dots,\tau$
\Beq \label{gen:temp:diff} 
    \begin{split} 
        d^n_i(\Aparam,\Cparam) & = \left( r_i + \dots + \gamma^{n-1} r_{i+n-1} + \gamma^{n} V(\state_{i+n};\Cparam) - V(\state_i;\Cparam) \right) \times \\  
        & \qquad \times \psi_b\!\left(\frac{\varphi(\bar\ctrl^n_i;\bar A(\bar\state^n_i;\Aparam)+\bar\eta^n_i, \Omega_2^n)} {\varphi(\bar\ctrl^n_i;\bar A^n_i+\bar\mu^n_i,\Omega_2^n)} \right), 
    \end{split} 
\Eeq
where $\psi_b$ is a~soft-truncating function, e.g. $\psi_b(x) = b\tanh(x/b)$, for a certain $b>1$.
\item Actor and Critic are updated. The improvement directions for Actor and Critic are 
\Beqs
    \Delta\Aparam & = \frac1\tau \sum_{n=1}^{\tau} \nabla_\Aparam \ln \varphi(\bar\ctrl^{n}_i;\bar A(\bar\state^{n}_i;\Aparam)+\bar\eta^n_i,\Omega_2^n) d^n_i(\Aparam,\Cparam) 
    - \nabla_\Aparam L(\state_i,\Aparam) 
    \label{Actor:adj} \\ 
    \Delta\Cparam & = \frac1\tau \sum_{n=1}^{\tau} \nabla_\Cparam V(\state_i;\Cparam) d^n_i(\Aparam,\Cparam), \label{Critic:adj} 
\Eeqs
where $L(\state,\Aparam)$ is a~loss function that penalizes Actor for producing actions that do not satisfy conditions e.g., they exceed their boundaries. $\Delta\Aparam$ is designed do increase/decrease the likelihood of the sequence of actions $\bar\ctrl^n_i$ proportionally to $d^n_i(\Aparam,\Cparam)$. $\Delta\Cparam$ is designed to make $V(\cdot\,;\Cparam)$ approximate the value function~\eqref{def:V} better. The improvement directions $\Delta\Aparam$ and $\Delta\Cparam$ are applied to update $\Aparam$ and $\Cparam$, respectively, with the use of either ADAM, SGD, or other method of stochastic optimization. 
\end{enumerate} 

In Point 1 the algorithm selects an~experienced event to replay. In Points~2 and 3 it determines the parameters the distribution of the sequence of subsequent actions, $\bar\ctrl^n_i$. In Point 4 it determines the relative quality of~$\bar\ctrl^n_i$. The temporal difference~\eqref{gen:temp:diff} implements two ideas. Firstly, $\Aparam$ is changing due to being optimized, thus the conditional distribution $(\bar\ctrl^n_i|\xi_{i-1})$ is now different than it was at the time when the actions $\bar\ctrl^n_i$ were happening. The density ratio in~\eqref{gen:temp:diff} accounts for this discrepancy of distributions. Secondly, in order to limit variance of the density ratio, the soft-truncating function~$\psi_b$ is applied. In Point 5 the parameters of Actor, $\Aparam$, and Critic, $\Cparam$, are being updated.

\section{Empirical study} 
\label{sec:experiments} 

This section presents simulations whose purpose has been to compare the algorithm introduced in Sec.~\ref{sec:alg} to state-of-the-art reinforcement learning methods. 
We compared the new algorithm (ACERAC) to ACER \cite{2009wawrzynski}, SAC \cite{2018haarnoja+3} and PPO \cite{2017schulman+4}. We used the rllib implementation \cite{2018liang} of SAC and PPO in the simulations. Our implementation of ACERAC is available at \url{https://github.com/mszulc913/acerac}.

\begin{figure}
    \centering
    \begin{tabular}{c c}
    \includegraphics[width=0.47\linewidth]{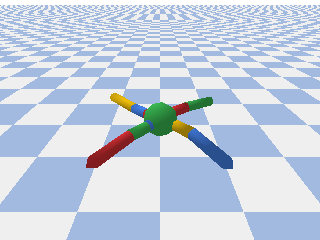} &
    \includegraphics[width=0.47\linewidth]{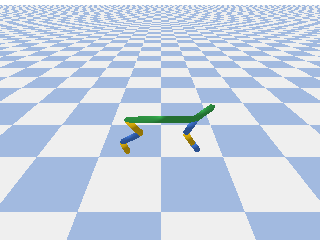} \\
    \includegraphics[width=0.47\linewidth]{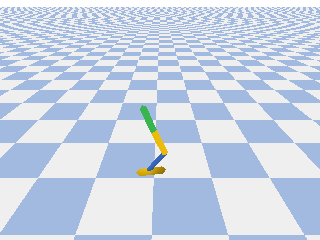} &
    \includegraphics[width=0.47\linewidth]{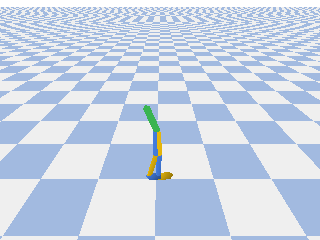}
    \end{tabular}
    \caption{Environments used in simulations:  Ant (left upper), HalfCheetah (right upper), Hopper (left lower), Walker2D (right lower).}
    \label{fig:Envs}
\end{figure}

We used four control tasks, namely Ant, Hopper, HalfCheetah, and Walker2D (see Fig.~\ref{fig:Envs}) from PyBullet physics simulator \cite{2019coumans} to compare the algorithms. A~simulator that is more popular in the RL community is MuJoCo \cite{2012todorov}.\footnote{We chose PyBullet because it is a~freeware, while MuJoCo is a~commercial software.} Hyperparameters that assure optimal performance of ACER, SAC, and PPO applied to the considered environments in MuJoCo are well known. However, PyBullet environments introduce several changes to MuJoCo tasks, which make them more realistic, thus more difficult. Additionally, physics in MuJoCo and PyBullets slightly differ \cite{2015erez+2}, hence we needed to tune the hyperparameters. Their value can be found in appendix \ref{AlgorithmsHyperparams}. 

For each learning algorithm we used Actor and Critic structures as described in \cite{2018haarnoja+3}. That is, both structures had the form of neural networks with two hidden layers of 256 units each.

\subsection{Experimental setting} 

Each learning run lasted for 3 million timesteps. Every 30000 timesteps of a~simulation was made with frozen weights and without exploration for 5 test episodes. An average sum of rewards within a~test episode was registered. Each run was repeated 5 times. 

\comment{
\begin{figure}
    \centering
    \begin{tabular}{c c}
    \includegraphics[width=0.47\linewidth]{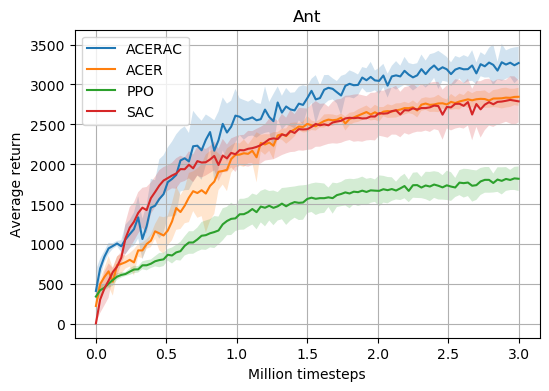} &
    \includegraphics[width=0.47\linewidth]{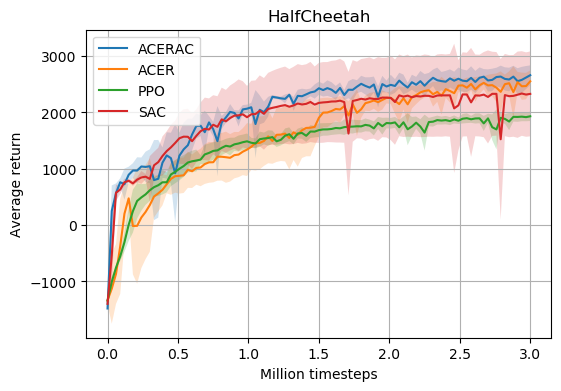} \\
    \includegraphics[width=0.47\linewidth]{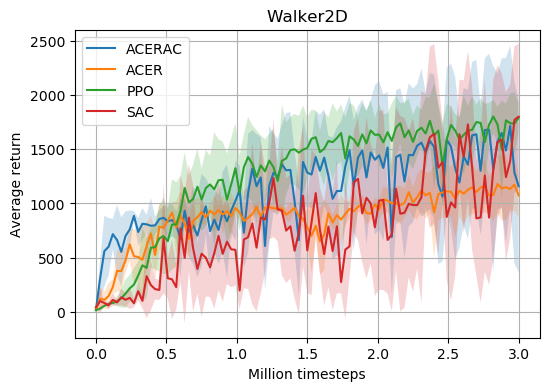}
    \end{tabular}
    \caption{ECML version. Learning curves for Ant (left upper), HalfCheetah (right upper) and Walker2D (left lower) environments: Average sums of rewards in test trials }
    \label{fig:Curves}
\end{figure}
}

\begin{figure}
    \centering
    \begin{tabular}{c c}
    \includegraphics[width=0.47\linewidth]{fig_AntBulletEnv-v0.png} &
    \includegraphics[width=0.47\linewidth]{fig_HalfCheetahBulletEnv-v0.png} \\
    \includegraphics[width=0.47\linewidth]{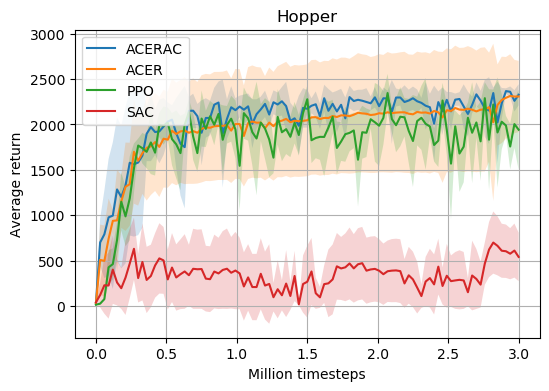} &
    \includegraphics[width=0.47\linewidth]{fig_Walker2DBulletEnv-v0.png}
    \end{tabular}
    \caption{Learning curves for Ant (left upper), HalfCheetah (right upper), Hopper (left lower) and Walker2D (right lower) environments: Average sums of rewards in test trials. }
    \label{fig:Curves}
\end{figure}

\subsection{Results} 

Figure~\ref{fig:Curves} presents learning curves for all four environments and all four compared algorithms. Each graph reports how a~sum of rewards in test episodes evolves within learning. Solid lines represent the average sums of rewards and shaded areas represent their standard deviations. 

It is seen that for Ant the algorithm that achieve the best performance is ACERAC, then ACER and SAC, then PPO. For HalfCHeetah, the best performance is achieved by ACERAC which is slightly better than ACER, then SAC, then PPO. For Hopper the algorithms to win are ACERAC {\it ex aequo} with ACER, then PPO, then SAC; actually SAC fails in this task. Eventually, for Walker2D, PPO achieves the best performance, then ACERAC and SAC, and then ACER. 

\subsection{Discussion} 

It is seen in Fig.~\ref{fig:Curves} that ACERAC is the best performing algorithm for three environments out of four (in one ACER preforms equally well). ACERAC extends ACER in two directions. Firstly, it admits autocrrelated actions. This enables exploration distributed in many actions instead in one. Secondly, in order to mimic learning with eligibility traces~\cite{1998kimura+2}, ACER estimates improvement directions with the use of a~sum whose limit is random. This increases variance of these estimates. Instead, for each state ACERAC computes an~improvement direction as an~average of increments similar to those ACER selects on random. Hence smaller variance of improvement direction estimates in ACERAC which enables larger step-sizes and faster learning. 

It is important to note that the algorithm introduced here, ACERAC, has been designed for fine time discretization and real life control problems. However, here it has been tested on simulated benchmarks in which time discretization was not particularly fine and control could be arbitrarily discontinuous. Its relatively good performance is a~desirable result. It allows to expect that this algorithm will perform relatively even better in real life control problems. That remains to be confirmed experimentally in further studies. 

\section{Conclusions and future work} 
\label{sec:conclusions} 

In this paper a~framework was introduced to apply reinforcement learning to policies that admit stochastic dependence between subsequent actions beyond state transition. This dependence is a~tool to enable reinforcement learning in physical systems and fine time discretization. It can also yield better exploration thus faster learning. 

An algorithm based on this framework, Actor-Critic with Experience Replay and Autocorrelated aCtions (ACERAC), was introduced. Its efficiency was verified in simulations of four learning control problems: Ant, HalfCheetah, Hopper, and Walker2D. The algorithm was compared with PPO, SAC, and ACER. ACERAC outperformed the competitors in Ant and HalfCheetah. For Hopper ACERAC was the best {\it ex aequo} with ACER. For Walker2D the best results was obtained by PPO. 

It is desirable to combine the framework proposed here with adaptation of dispersion of actions by introducing reward for the entropy of their distribution, as it is done in PPO. The framework proposed here is specially designed for applications in robotics. An obvious step of our further research is to apply it in this field, obviously much more demanding than simulations. 

\section*{Acknowledgement}

This work was partially funded by a grant of Warsaw University of Technology Scientific Discipline Council for Computer Science and Telecommunications. 


\begin{thebibliography}{10}
\providecommand{\url}[1]{\texttt{#1}}
\providecommand{\urlprefix}{URL }
\providecommand{\doi}[1]{https://doi.org/#1}

\bibitem{1983barto+2}
Barto, A.G., Sutton, R.S., Anderson, C.W.: Neuronlike adaptive elements that
  can learn difficult learning control problems. IEEE Transactions on Systems,
  Man, and Cybernetics B  \textbf{13},  834--846 (1983)

\bibitem{2019coumans}
Coumans, E., Bai, Y.: Pybullet, a python module for physics simulation for
  games, robotics and machine learning. \url{http://pybullet.org} (2016--2019)

\bibitem{2015erez+2}
{Erez}, T., {Tassa}, Y., {Todorov}, E.: Simulation tools for model-based
  robotics: Comparison of bullet, havok, mujoco, ode and physx. In: 2015 IEEE
  International Conference on Robotics and Automation (ICRA). pp. 4397--4404
  (2015)

\bibitem{2018haarnoja+3}
Haarnoja, T., Zhou, A., Abbeel, P., Levine, S.: Soft actor-critic: Offpolicy
  maximum entropy deep reinforcement learning with a stochastic actor (2018),
  arXiv:1801.01290

\bibitem{2017vanhoof+2}
van Hoof, H., Tanneberg, D., Peters, J.: Generalized exploration in policy
  search. Machine Learning  \textbf{106},  1705--1724 (2017)

\bibitem{2002kakade+1}
Kakade, S., Langford, J.: Approximately optimal approximate reinforcement
  learning. In: Proceedings of the Nineteenth International Conference on
  Machine Learning, ICML’02. pp. 267--274 (2002)

\bibitem{1998kimura+2}
Kimura, H., Kobayashi, S.: An analysis of actor/critic algorithms using
  eligibility traces: Reinforcement learning with imperfect value function. In:
  ICML (1998)

\bibitem{2019korenkevych+3}
Korenkevych, D., Mahmood, A.R., Vasan, G., Bergstra, J.: Autoregressive
  policies for continuous control deep reinforcement learning. In: Proceedings
  of the Twenty-Eighth International Joint Conference on Artificial
  Intelligence (IJCAI-19). pp. 2754--2762 (2019)

\bibitem{2018liang}
Liang, E., Liaw, R., Nishihara, R., Moritz, P., Fox, R., Goldberg, K.,
  Gonzalez, J., Jordan, M., Stoica, I.: {RL}lib: Abstractions for distributed
  reinforcement learning. In: Dy, J., Krause, A. (eds.) Proceedings of the 35th
  International Conference on Machine Learning. Proceedings of Machine Learning
  Research, vol.~80, pp. 3053--3062. PMLR, Stockholmsmässan, Stockholm Sweden
  (10--15 Jul 2018)

\bibitem{2016lillicrap+7}
Lillicrap, T.P., Hunt, J.J., Pritzel, A., Heess, N., Erez, T., Tassa, Y.,
  Silver, D., Wierstra, D.: Continuous control with deep reinforcement learning
  (2016), arXiv:1509.02971

\bibitem{2013mnih+6}
Mnih, V., Kavukcuoglu, K., Silver, D., Graves, A., Antonoglou, I., Wierstra,
  D., Riedmiller, M.: Playing atari with deep reinforcement learning (2013),
  arXiv:1312.5602

\bibitem{2015schulman+4}
Schulman, J., Levine, S., Moritz, P., Jordan, M.I., Abbeel, P.: Trust region
  policy optimization (2015), arXiv:1502.05477

\bibitem{2017schulman+4}
Schulman, J., Wolski, F., Dhariwal, P., Radford, A., Klimov, O.: Proximal
  policy optimization algorithms (2017), arXiv:1707.06347

\bibitem{2012todorov}
Todorov, E., Erez, T., Tassa, Y.: Mujoco: A physics engine for model-based
  control. In: 2012 IEEE/RSJ International Conference on Intelligent Robots and
  Systems. pp. 5026--5033. IEEE (2012)

\bibitem{2016wang+6}
Wang, Z., Bapst, V., Heess, N., Mnih, V., Munos, R., Kavukcuoglu, K.,
  de~Freitas, N.: Sample efficient actor-critic with experience replay (2016),
  arXiv:1611.01224

\bibitem{2007wawrzynski}
Wawrzyński, P.: Learning to control a 6-degree-of-freedom walking robot. In:
  Proceedings of EUROCON 2007 The International Conference on Computer as a
  Tool. pp. 698--705 (2007)

\bibitem{2009wawrzynski}
Wawrzyński, P.: Real-time reinforcement learning by sequential actor–critics
  and experience replay. Neural Networks  \textbf{22}(10),  1484--1497 (2009)

\bibitem{2015wawrzynski}
Wawrzyński, P.: Control policy with autocorrelated noise in reinforcement
  learning for robotics. International Journal of Machine Learning and
  Computing  \textbf{5}(2),  91--95 (2015)

\end{thebibliography}

\clearpage 

\appendix 

\section{Algorithms' hyperparameters} 
\label{AlgorithmsHyperparams}

This section presents hyperparameters used in simulations reported in Sec.~\ref{sec:experiments}. All algorithms used the discount factor equal to $0.99$. The rest of hyperparameters for ACERAC, ACER, SAC, and PPO, are depicted in Tab.~\ref{tab:ACERACParams}, \ref{tab:ACERParams}, \ref{tab:SACParams}, and \ref{tab:PPOParams}, respectively. 

\begin{multicols}{2} 

    \noindent
    \parbox{\linewidth}{
    \centering
    \begin{tabular}{c|c}
        \hline
        Parameter & Value \\
        \hline
        Action std. dev. for Hopper & 0.3 \\
        Action std. dev. for other envs. & 0.4 \\
        $\alpha$ & 0.5 \\ 
        Critic step-size for Walker2D & $10^{-4}$ \\
        Critic step-size for other envs. & $6\cdot10^{-5}$ \\
        Actor step-size for Walker2D & $5\cdot10^{-5}$ \\
        Actor step-size for other envs. & $3\cdot10^{-5}$ \\
        $\tau$ & 4 \\
        $b$ & 2 \\
        Memory size & $10^6$\\
        Minibatch size & 256 \\
        Target update interval & 1 \\
        Gradient steps & 1 \\
        Learning start & $10^3$ \\
        \hline
    \end{tabular}
    \captionof{table}{ACERAC hyperparameters}
    \label{tab:ACERACParams}
    }
    \vspace{2em}

    \noindent
    \parbox{\linewidth}{
    \centering
    \begin{tabular}{c|c}
        \hline
        Parameter & Value \\
        \hline
        Action std. dev. & 0.3 \\
        Critic step-size & $10^{-5}$ \\
        Actor step-size & $10^{-5}$ \\ 
        $\lambda$ & 0.9 \\
        $b$ & 2 \\
        Memory size & $10^6$\\
        Minibatch size & 256 \\
        Target update interval & 1 \\
        Gradient steps & 1 \\
        Learning start & $10^3$ \\
        \hline
    \end{tabular}
    \captionof{table}{ACER hyperparameters}
    \label{tab:ACERParams}
    }
    \vspace{2em}

    \noindent
    \parbox{\linewidth}{
    \centering
    \begin{tabular}{c|c}
        \hline
        Parameter & Value \\
        \hline
        Step-size for Hopper & 0.0001 \\
        Step-size for other envs & 0.0003 \\
        Replay buffer size & $10^6$ \\
        Minibatch size & 256 \\
        Target smoothing coef. $\tau$ & 0.005 \\
        Target update interval & 1 \\
        Gradient steps & 1 \\
        Learning start for Ant & $10^4$ \\
        Learning start for HalfCheetah & $10^4$ \\
        Learning start for Hopper & $10^3$ \\
        Learning start for Walker2D & $10^3$ \\
        Reward scale for Ant & 1 \\
        Reward scale for HalfCheetah & 0.1 \\
        Reward scale for Hopper & 30 \\
        Reward scale for Walker2D & 30 \\
        \hline
    \end{tabular}
    \captionof{table}{SAC hyperparameters}
    \label{tab:SACParams}
    }
    \vspace{2em} 


    \noindent
    \parbox{\linewidth}{
    \centering
    \begin{tabular}{c|c}
        \hline
        Parameter & Value \\
        \hline
        GAE parameter ($\lambda$) & 0.95 \\
        Minibatch size & 64 \\
        Step-size & 0.0003 \\
        Horizon & 2048 \\
        Number of epochs & 10 \\
        Policy clipping coef. & 0.2 \\
        Value function clipping coef. & 10 \\
        Target KL & 0.01 \\
        \hline
    \end{tabular}
    \captionof{table}{PPO hyperparameters}
    \label{tab:PPOParams}
    }

\end{multicols} 

\end{document}